\documentclass[letterpaper, journal, doublecolumn]{IEEEtran}
\usepackage{cite}
\usepackage{amsmath,amssymb,amsfonts}
\usepackage{algorithmic}
\usepackage{graphicx}
\usepackage{textcomp}
\usepackage{tabularx}  
\usepackage{ragged2e}  
\usepackage{makecell}  
\usepackage{array} 
\usepackage{mdframed}
\usepackage{comment}
\usepackage{url}
\usepackage[T1]{fontenc}
\usepackage{csquotes}
\usepackage{float}
\usepackage{tabularx}  
\usepackage{booktabs}  
\usepackage{multirow}  

\newcolumntype{L}{>{\RaggedRight\arraybackslash}X} 

\def\BibTeX{{\rm B\kern-.05em{\sc i\kern-.025em b}\kern-.08em
    T\kern-.1667em\lower.7ex\hbox{E}\kern-.125emX}}

\title{\LARGE \bf
Bias in Emotion Recognition with ChatGPT
}

\author{
Naoki Wake$^{1}$,
Atsushi Kanehira$^{1}$,
Kazuhiro Sasabuchi$^{1}$,
Jun Takamatsu$^{1}$,
and Katsushi Ikeuchi$^{1}$
\thanks{$^{1}$Applied Robotics Research, Microsoft, 
        Redmond, WA 98052, USA
        {\tt\small naoki.wake@microsoft.com}}%
}

\begin{document}
\maketitle
\thispagestyle{empty}
\pagestyle{empty}
\begin{abstract}
This technical report explores the ability of ChatGPT in recognizing emotions from text, which can be the basis of various applications like interactive chatbots, data annotation, and mental health analysis. While prior research has shown ChatGPT's basic ability in sentiment analysis, its performance in more nuanced emotion recognition is not yet explored. Here, we conducted experiments to evaluate its performance of emotion recognition across different datasets and emotion labels. Our findings indicate a reasonable level of reproducibility in its performance, with noticeable improvement through fine-tuning. However, the performance varies with different emotion labels and datasets, highlighting an inherent instability and possible bias. The choice of dataset and emotion labels significantly impacts ChatGPT's emotion recognition performance. This paper sheds light on the importance of dataset and label selection, and the potential of fine-tuning in enhancing ChatGPT's emotion recognition capabilities, providing a groundwork for better integration of emotion analysis in applications using ChatGPT. 
\end{abstract}

\section{Introduction}
\label{sec:introduction}
The ability to recognize emotions from text is a fundamental aspect of various applications ranging from interactive chatbots\cite{terada2012artificial,loffler2018multimodal}, annotated data generation\cite{koptyra2023clarin}, text analysis\cite{zhang2023clusterllm}, and mental health analysis\cite{yang2023evaluations,lamichhane2023evaluation,amin2023will}. As chatbots become increasingly sophisticated, understanding the emotional nuance within user inputs can significantly enhance the conversational experience by enabling more empathetic and contextually appropriate responses. Similarly, in the realm of mental health, analyzing textual data for emotional content can provide crucial insights into an individual's mental health conditions.

In recent years, the advancements in Large Language Model (LLM), notably models like ChatGPT, have shown promise in various Natural Language Processing (NLP) tasks, including emotion recognition. Early studies have demonstrated the capability of ChatGPT in basic sentiment analysis tasks where the primary goal is to distinguish between positive and negative sentiments\cite{amin2023will,qin2023chatgpt,zhong2023can,kocon2023chatgpt}. Beyond this binary classification a few reports delved into a nuanced understanding of emotions like joy, sadness, anger, and surprise and showed that ChatGPT shows reasonable performance in such detailed emotion analysis\cite{zhao2023chatgpt,yang2023evaluations} in a zero-shot and few-shot prompting conditions.

However, one of the caveats associated with LLMs is their tendency to exhibit biases in the generated outputs. These biases often have roots in the non-uniformity of their training data and the labeling methodologies employed\cite{sohail2023decoding}. Consequently, it is likely that potential biases manifest in the nuances of emotional recognition. While understanding these specific tendencies in emotion recognition could offer insights for the responsible deployment of ChatGPT in real-world applications, detailed studies on this particular aspect remain limited.

Using ChatGPT as a representative LLM, this study delves into its emotion recognition capabilities across various datasets and emotion labels. In experiments that solely relied on prompting (i.e., a zero-shot condition), ChatGPT exhibited commendable emotion recognition performance, consistent with previous research findings. However, we observed disparities in performance across different emotion labels, and intriguingly, the same label sometimes yielded varied results across datasets. While fine-tuning generally enhanced the model's recognition abilities, inconsistencies among labels persisted. A notable observation was the sensitivity of the model to slight label name changes (e.g., from ``happiness'' to ``happy''). Despite these fluctuations, ChatGPT's responses remained largely consistent, and the labeling context had minimal influence on its overall performance. Our findings not only elucidate ChatGPT’s strengths and limitations in emotion recognition but also set the stage for future investigations focused on optimizing performance and minimizing biases in ChatGPT-driven emotion recognition tasks.


\section{Methods}
\begin{table*}[htbp]
\caption{The detailed information of datasets for emotion recognition}
\label{table:dataset}
\centering
\begin{tabular}{llcl}
\hline
\textbf{Dataset} & \textbf{Data Source} & \textbf{Conv./Utter.} & \textbf{Emotion Label} \\
\hline
IEMOCAP & Multimodal Act (only text modality) & 31/1,622 & neutral, sad, anger, happy, frustrated, excited \\
MELD & Friends TV Show & 280/2,610 & neutral, sadness, anger, disgust, fear, joy, surprise \\
EmoryNLP & Friends TV Show  & 85/1,328 & neutral, sad, mad, scared, powerful, peaceful, joyful \\
DailyDialog & Human Written Scripts & 1,000/7,740 & neutral, happiness, surprise, sadness, anger, disgust, fear \\
\hline
\end{tabular}
\end{table*}
\subsection{ChatGPT and prompts}
ChatGPT is an LLM developed by OpenAI\cite{OpenAI}. One of its distinct features is its capability to generate text in a conversational manner. This interactive nature allows users to leverage the model for a wide range of NLP tasks using conversational texts, which is called a prompt. The methodology of crafting prompts to obtain desired outputs from the off-the-shelf model without additional dataset (i.e, zero-shot setting) is referred to as prompt engineering~\cite{kocon2023chatgpt}. Additionally, there is a methodology that employs a dataset of expected input-output pairs to fine-tune the model's behavior. In this research, we employed the gpt-3.5-turbo, a fixed version of the GPT model that is publicly available for fine-tuning. This version represents the latest iteration of ChatGPT that publicly supports fine-tuning to date. We assessed its efficacy in both prompt engineering and fine-tuning conditions.


We evaluate ChatGPT's ability to discern emotions in complex scenarios based on the emotion recognition in conversation (ERC) task\cite{poria2019emotion}. In ERC, the goal is to recognize the emotion of each utterance within a conversation from a predefined set of emotion categories. In this paper, we assume an ongoing conversation scenario where future conversation information is unobserved. We estimate the emotion label for the current utterance based solely on the history of past conversations. That is, given a conversation history \( C = \{u_1, u_2, \ldots, u_t\} \) consisting of a sequence of \( t \) utterances, the task is to recognize the emotion label to \(u_t\), referred to as \(y_t\). Here, \( y_t \in E \), where \( E \) is the set of predefined emotion labels.

For the zero-shot setting, we designed a prompt based on a previous study\cite{yang2023evaluations}:
\blockquote{\textit{You are an emotion analyzer capable of understanding the sentiment within a text. Previous dialogue (sequence of pairs of speaker and the utterance): ``[Dialogue History]''. (Your task) Consider this dialogue to assign one emotion label to the next utterance: ``[Query]''. Only one label. Only from this emotion list: [Label List].}}, 
where [Dialogue History] denotes the history of dialogue containing a list of speaker names and utterances, [Query] denotes a target utterance with the speaker name, and [Label List] denotes emotion labels for a dataset. When using a dataset without speaker IDs, the name was denoted as `unknown.' 

\subsection{Dataset and Evaluation Metrics}
We evaluates the performance of emotion recognition by ChatGPT across five datasets: IEMOCAP\cite{busso2008iemocap}, MELD\cite{poria2018meld}, EmoryNLP\cite{zahiri2018emotion}, DailyDialog\cite{li2017dailydialog}. For the IEMOCAP multimodal dataset, we focus solely on the text modality. The detailed information of those dataset is summarized in Table~\ref{table:dataset}. 

In our experiments, we adhered to the established metrics in line with the prevailing approach in the literature. Specifically, for the IEMOCAP, MELD, and EmoryNLP datasets, we employed the Weighted-F1 metric. For the DailyDialog dataset, considering the disproportionately high occurrence of neutral utterances, we utilized the micro-F1 metric, excluding the neutral category. It is noteworthy that our analysis was not limited to overall performance but also we also examined the recognition performance for each individual emotion label, to explore the specific tendencies in emotion recognition.

\subsection{Fine-tuning}
\begin{figure}[ht]
  \centering
  \includegraphics[width=0.3\textwidth]{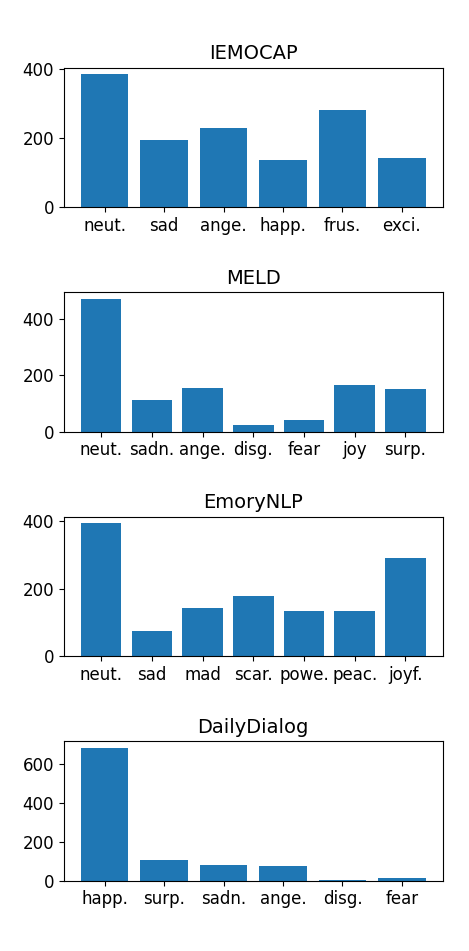}
  \caption{
  The distribution of labels used for fine-tuning ChatGPT. For the DailyDialog dataset, we eliminated the neutral labels due to its occurrence.
  }
  \label{fig:label_distribution}
\end{figure}

When fine-tuning ChatGPT, we used the prompts from the zero-shot experiments along with their corresponding correct labels. To test with a more generalized model, we combined four distinct datasets previously mentioned for fine-tuning, and tested the model across the four datasets. We utilized data from the validation sets of each dataset: 1365 samples from IEMOCAP, 1109 from MELD, 1344 from EmoryNLP, and 961 from DailyDialog. The distribution of the labels are shown in Fig.\ref{fig:label_distribution}

\section{Results}
\subsection{Main Results}
The overall performance in emotion recognition significantly exceeded chance levels, aligning with the findings from previous studies (Table\ref{tab:emotion_metrics_IEMOCAP},\ref{tab:emotion_metrics_MELD},\ref{tab:emotion_metrics_EmoryNLP},\ref{tab:emotion_metrics_DailyDialog}). However, there was noticeable variance when examining performance based on individual emotion labels. For example, in the zero-shot experiments on the EmoryNLP dataset, the recognition performance (F1 score) for `joyful' was 51\%, while it was below 5\% for `peaceful' and `powerful.'

\begin{figure}[ht]
  \centering
  \includegraphics[width=0.48\textwidth]{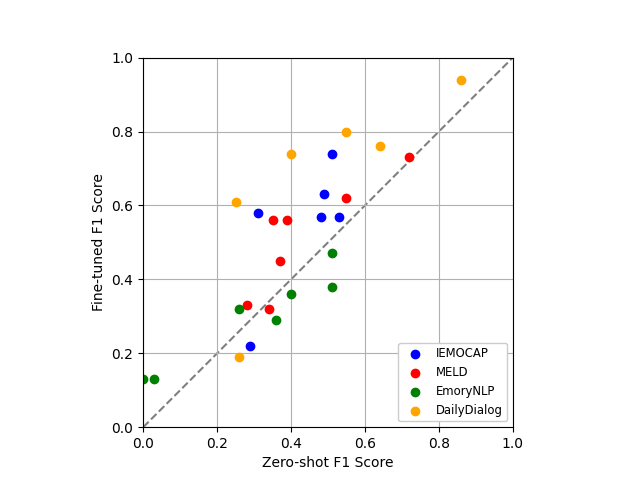}
  \caption{
  Comparison of emotion recognition performance between zero-shot and fine-tuned models across four datasets.
  }
  \label{fig:label_plot1}
\end{figure}

Fine-tuning generally observed to enhance performance (Fig.\ref{fig:label_plot1}). Interestingly, despite the EmoryNLP dataset using text based on the same TV series as MELD, the impact of fine-tuning wasn't as pronounced. 

Despite the variability in the labels of the datasets used for fine-tuning, a strong correlation was observed between the performance in zero-shot and fine-tuning (correlation coefficient of 0.81). This result suggests that while fine-tuning improves performance, it still retains inherent tendencies.

\begin{figure}[ht]
  \centering
  \includegraphics[width=0.48\textwidth]{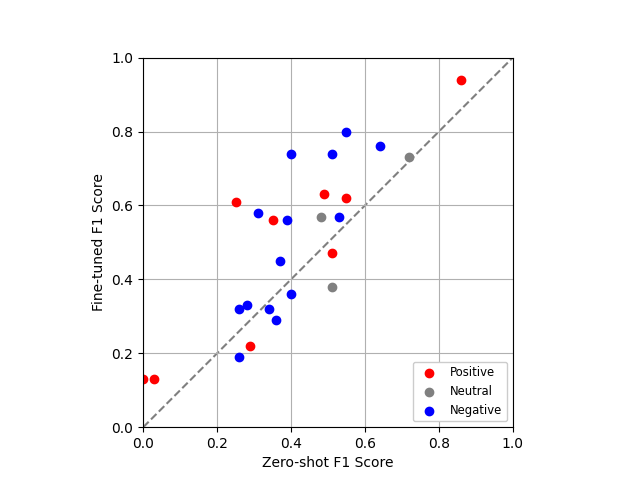}
  \caption{
  Comparison of emotion recognition performance between zero-shot and fine-tuned models colored with positive/neutral/negative labels.
  }
  \label{fig:label_plot2}
\end{figure}

To explore the potential semantic relationship between recognition performance and label tendencies, we classified the labels into three overarching categories: positive, neutral, and negative (Fig.\ref{fig:label_plot2}). However, our analysis did not reveal any discernible correlation between recognition performance and the semantic nature of the labels.

\subsection{Sensitivity to Label Text}
From these results, we observed that recognition performance varied significantly across datasets, even for similar labels. For instance, in the DailyDialog dataset, the recognition performance for 'happiness' was around 94\%, while in the IEMOCAP dataset, the recognition performance for 'happy' was only about 22\% (see the fine-tuning columns in Table\ref{tab:emotion_metrics_IEMOCAP} and \ref{tab:emotion_metrics_DailyDialog}). Notably, this tendency is observed in the zero-shot condition, as well. While these differences may be influenced by the unique characteristics of the sentences contained within each dataset, we hypothesized that there might be sensitivity to the label text. To test this, we changed the label name in the IEMOCAP dataset from `happy' to `happiness' and applied the same model to evaluate its performance.

As a result, the recognition performance for `happiness' improved to 43\% (Table\ref{tab:emotion_metrics_IEMOCAP_happiness}). Additionally, the recognition performance for other labels did not suffer significantly, indicating that changing the label improved the recognition performance for the `happy' label without compromising the performance of other labels. This result suggests a sensitivity of the language model to the specific text of the labels.
\subsection{Reproducibility of the Performance}

As an additional experiment, we investigated the reproducibility of ChatGPT's results using Cohen's Kappa score \cite{cohen1960coefficient}. Under the zero-shot condition, when applying the ChatGPT model twice to the MELD test data, the Kappa score was 0.65. This suggests that we can expect reasonable reproducibility when applying the model multiple times.

\subsection{Dataset selection}
In the experiment, the model was fine-tuned on a blend of all datasets. This dataset mixture can prevent over-fitting to a specific dataset, but it also has the potential to significantly affect the model performance. To gain insights on the impact of dataset mixture on the main result, we compared the results of fine-tuning on a single dataset against those trained on the mixed dataset, using the MELD dataset as an example. The model trained on a single dataset performed better for most labels, which is reasonable. However, the difference was at most about 7 percent, which is marginal compared to the difference across labels (49 percent at most), and there was no significant difference in the trend of performance differences between labels. These results suggest that the model's recognition bias is less likely to be affected by the scope of the dataset.

\begin{table}[H]
\centering
\caption{Emotion Recognition Metrics (IEMOCAP)}
\resizebox{\columnwidth}{!}{
\label{tab:emotion_metrics_IEMOCAP}
\begin{tabular}{l*{3}{c}*{3}{c}}
\toprule
\multirow{2}{*}{Emotion} & \multicolumn{3}{c}{zero-shot} & \multicolumn{3}{c}{fine-tuning} \\
 & Precision & Recall & F1 & Precision & Recall & F1 \\
\midrule
neutral & 0.36 & 0.70 & 0.48 & 0.67 & 0.49 & \textbf{0.57} \\
sad & 0.79 & 0.38 & 0.51 & 0.85 & 0.65 & \textbf{0.74} \\
anger & 0.70 & 0.20 & 0.31 & 0.47 & 0.74 & \textbf{0.58} \\
happy & 0.33 & 0.25 & \textbf{0.29} & 0.24 & 0.20 & 0.22 \\
frustrated & 0.52 & 0.54 & 0.53 & 0.53 & 0.62 & \textbf{0.57} \\
excited & 0.65 & 0.39 & 0.49 & 0.62 & 0.64 & \textbf{0.63} \\
\addlinespace
macro-average & 0.56 & 0.41 & 0.43 & 0.56 & 0.56 & \textbf{0.55} \\
std & 0.19 & 0.19 & 0.11 & 0.21 & 0.19 & 0.17 \\
weighted-f1 (Ours) & -- & -- & 0.46 & -- & -- & \textbf{0.58} \\
\midrule
weighted-f1 \cite{zhao2023chatgpt} & -- & -- & 0.45 & -- & -- & -- \\
\midrule
weighted-f1 \cite{yang2023evaluations} & -- & -- & 0.53 & -- & -- & -- \\
\bottomrule
\end{tabular}
}
\begin{flushright}
\footnotesize Bold denotes higher values between zero-shot and fine-tuning.
\end{flushright}
\end{table}

\begin{table}[H]
\centering
\caption{Emotion Recognition Metrics (MELD)}
\resizebox{\columnwidth}{!}{
\label{tab:emotion_metrics_MELD}
\begin{tabular}{l*{3}{c}*{3}{c}}
\toprule
\multirow{2}{*}{Emotion} & \multicolumn{3}{c}{zero-shot} & \multicolumn{3}{c}{fine-tuning} \\
 & Precision & Recall & F1 & Precision & Recall & F1 \\
\midrule
neutral & 0.64 & 0.84 & 0.72 & 0.79 & 0.67 & \textbf{0.73} \\
sadness & 0.48 & 0.31 & 0.37 & 0.43 & 0.48 & \textbf{0.45} \\
anger & 0.65 & 0.28 & 0.39 & 0.53 & 0.59 & \textbf{0.56} \\
disgust & 0.33 & 0.34 & \textbf{0.34} & 0.37 & 0.28 & 0.32 \\
fear & 0.22 & 0.38 & 0.28 & 0.33 & 0.33 & \textbf{0.33} \\
joy & 0.55 & 0.55 & 0.55 & 0.61 & 0.62 & \textbf{0.62} \\
surprise & 0.58 & 0.25 & 0.35 & 0.47 & 0.69 & \textbf{0.56} \\
\addlinespace
macro-average & 0.49 & 0.42 & 0.43 & 0.50 & 0.52 & \textbf{0.51} \\
std & 0.16 & 0.21 & 0.15 & 0.16 & 0.17 & 0.15 \\
weighted-f1 (Ours) & -- & -- & 0.57 & -- & -- & \textbf{0.63} \\
\midrule
weighted-f1 \cite{zhao2023chatgpt} & -- & -- & 0.57 & -- & -- & -- \\
\midrule
weighted-f1 \cite{yang2023evaluations} & -- & -- & 0.61 & -- & -- & -- \\
\bottomrule
\end{tabular}
}
\end{table}

\begin{table}[H]
\centering
\caption{Emotion Recognition Metrics (EmoryNLP)}
\resizebox{\columnwidth}{!}{
\label{tab:emotion_metrics_EmoryNLP}
\begin{tabular}{l*{3}{c}*{3}{c}}
\toprule
\multirow{2}{*}{Emotion} & \multicolumn{3}{c}{zero-shot} & \multicolumn{3}{c}{fine-tuning} \\
 & Precision & Recall & F1 & Precision & Recall & F1 \\
\midrule
neutral & 0.37 & 0.81 & \textbf{0.51} & 0.42 & 0.35 & 0.38 \\
sad & 0.41 & 0.32 & \textbf{0.36} & 0.33 & 0.27 & 0.29 \\
mad & 0.42 & 0.38 & \textbf{0.40} & 0.34 & 0.38 & 0.36 \\
scared & 0.38 & 0.19 & 0.26 & 0.29 & 0.36 & \textbf{0.32} \\
powerful & 0.00 & 0.00 & 0.00 & 0.16 & 0.11 & \textbf{0.13} \\
peaceful & 0.22 & 0.02 & 0.03 & 0.17 & 0.10 & \textbf{0.13} \\
joyful & 0.54 & 0.48 & \textbf{0.51} & 0.39 & 0.57 & 0.47 \\
\addlinespace
macro-average & 0.34 & 0.31 & 0.29 & 0.30 & 0.31 & \textbf{0.30} \\
std & 0.17 & 0.28 & 0.21 & 0.10 & 0.16 & 0.13 \\
weighted-f1 (Ours) & -- & -- & \textbf{0.35} & -- & -- & 0.33 \\
\midrule
weighted-f1 \cite{zhao2023chatgpt} & -- & -- & 0.37 & -- & -- & -- \\
\midrule
weighted-f1 \cite{yang2023evaluations} & -- & -- & 0.33 & -- & -- & -- \\
\bottomrule
\end{tabular}
}
\end{table}

\begin{table}[H]
\centering
\caption{Emotion Recognition Metrics (DailyDialog)}
\resizebox{\columnwidth}{!}{
\label{tab:emotion_metrics_DailyDialog}
\begin{tabular}{l*{3}{c}*{3}{c}}
\toprule
\multirow{2}{*}{Emotion} & \multicolumn{3}{c}{zero-shot} & \multicolumn{3}{c}{fine-tuning} \\
 & Precision & Recall & F1 & Precision & Recall & F1 \\
\midrule
neutral & -- & -- & -- & -- & -- & -- \\
happiness & 0.97 & 0.77 & 0.86 & 0.98 & 0.90 & \textbf{0.94} \\
surprise & 0.53 & 0.16 & 0.25 & 0.48 & 0.85 & \textbf{0.61} \\
sadness & 0.57 & 0.54 & 0.55 & 0.74 & 0.87 & \textbf{0.80} \\
anger & 0.80 & 0.53 & 0.64 & 0.73 & 0.79 & \textbf{0.76} \\
disgust & 0.39 & 0.19 & \textbf{0.26} & 0.71 & 0.11 & 0.19 \\
fear & 0.26 & 0.82 & 0.40 & 0.72 & 0.76 & \textbf{0.74} \\
\addlinespace
macro-average & 0.59 & 0.50 & 0.49 & 0.73 & 0.71 & \textbf{0.67} \\
std & 0.26 & 0.28 & 0.24 & 0.16 & 0.30 & 0.26 \\
micro-F1 (Ours) & -- & -- & 0.75 & -- & -- & \textbf{0.86} \\
\midrule
micro-F1 \cite{zhao2023chatgpt} & -- & -- & 0.41 & -- & -- & -- \\
\midrule
micro-F1 \cite{yang2023evaluations} & -- & -- & 0.43 & -- & -- & -- \\
\bottomrule
\end{tabular}
}
\end{table}

\begin{table}[H]
\centering
\caption{Recognition Performance With an Altered Label (IEMOCAP) }
\label{tab:emotion_metrics_IEMOCAP_happiness}
\begin{tabular}{l*{3}{c}*{3}{c}}
\toprule
\multirow{2}{*}{Emotion} & \multicolumn{3}{c}{fine-tuning} \\
 & Precision & Recall & F1 \\
\midrule
neutral & 0.68 & 0.57 & 0.62 \\
sad & 0.86 & 0.68 & 0.76 \\
anger & 0.45 & 0.72 & 0.55 \\
\textbf{happiness} & 0.41 & 0.45 & 0.43 \\
frustrated & 0.58 & 0.64 & 0.61 \\
excited & 0.68 & 0.62 & 0.65 \\
\addlinespace
macro-average & 0.61 & 0.61 & 0.60 \\
std & 0.17 & 0.09 & 0.11\\
weighted-f1 & -- & -- & 0.62 \\
\bottomrule
\end{tabular}
\end{table}

\begin{table}[H]
\centering
\caption{Recognition Performance After Fine-Tuning on a Single Dataset (MELD)}
\resizebox{\columnwidth}{!}{
\label{tab:emotion_metrics_datasetchoice}
\begin{tabular}{l*{3}{c}*{3}{c}}
\toprule
\multirow{2}{*}{Emotion} & \multicolumn{3}{c}{fine-tuning on a single dataset} & \multicolumn{3}{c}{fine-tuning on a mixed dataset} \\
 & Precision & Recall & F1 & Precision & Recall & F1 \\
\midrule
neutral & 0.78 & 0.82 & \textbf{0.80} & 0.79 & 0.67 & 0.73 \\
sadness & 0.52 & 0.41 & \textbf{0.46} & 0.43 & 0.48 & 0.45 \\
anger & 0.61 & 0.64 & \textbf{0.62} & 0.53 & 0.59 & 0.56 \\
disgust & 0.52 & 0.22 & 0.31 & 0.37 & 0.28 & \textbf{0.32}\\
fear & 0.39 & 0.28 & \textbf{0.33} & 0.33 & 0.33 & \textbf{0.33} \\
joy & 0.62 & 0.65 & \textbf{0.64} & 0.61 & 0.62 & 0.62 \\
surprise & 0.60 & 0.57 & \textbf{0.59} & 0.47 & 0.69 & 0.56 \\
\addlinespace
macro-average & 0.58 & 0.51 & \textbf{0.53} & 0.50 & 0.52 & 0.51 \\
std & 0.12 & 0.22 & 0.18 & 0.16 & 0.17 & 0.15 \\
weighted-f1 (Ours) & -- & -- & \textbf{0.68} & -- & -- & 0.63 \\
\bottomrule
\end{tabular}
}
\end{table}
\section{Discussion}\label{discussion}
In this study, we evaluated ChatGPT's ability to recognize emotions associated with text. Within the context of the ERC task, ChatGPT demonstrated a commendable performance, surpassing chance levels and aligning with prior research findings. While fine-tuning generally improved recognition, we observed remaining variability in performance across different emotion labels and datasets.

The variability in recognition performance can be attributed to biases in the training data and the labeling methodologies used\cite{sohail2023decoding}. In prior research that attempted to generate datasets with emotion labels using ChatGPT, Koptyra et al. \cite{koptyra2023clarin} reported variability in the number of generated labels. This observation aligns with our result and both suggest that the inherent bias in ChatGPT may influence emotion analysis. In the context of emotion recognition, biases towards certain label classes have been reported in pioneering language models\cite{mao2022biases}. Consequently, it is of utmost importance for researchers to understand and account for the intrinsic biases of language models when applying them to emotion analysis.

LLMs are praised for their advanced linguistic skills, making them an appealing solution for various applications, including mental health (\cite{yang2023evaluations,lamichhane2023evaluation,amin2023will,elyoseph2023chatgpt,uludag2023testing}; see \cite{he2023survey} for the survey). Given the potential biases reflected in emotion recognition, it would be prudent for researchers and practitioners to exercise caution when applying applying these models in sensitive domains like mental health. Over-reliance on these models without thorough evaluation might inadvertently propagate biases or yield misleading results.

Even within datasets such as MELD and EmoryNLP, which are based on the same television series, there was a notable variance in performance, with particularly low performances for ``powerful'' and ``peaceful.'' We hypothesize that one reason these emotion labels are not adequately associated with the text might be due to biases within the knowledge base of ChatGPT. Figure \ref{fig:chatgpt_response} presents examples of ChatGPT's responses when inquired about commonly used emotion labels. The responses from ChatGPT encompass widely accepted emotional definitions, such as Ekman's basic emotions\cite{ekman1992there} and Plutchik's wheel of emotions\cite{plutchik1980general}. While the input-output mechanisms of LLM are black-box in nature, the fact that the model produces such knowledge suggests that for ChatGPT, associating these emotions with the input text might be relatively easier than associating less commonly used labels. These experimental results, alongside the sensitivity of the labels to the text, highlight the importance of label selection.

\subsection{Limitation of this Study}
While our study sheds light on several aspects of emotion recognition using ChatGPT, there are certain limitations that should be acknowledged:
\begin{enumerate}
    \item \textbf{Inconsistencies in labeling criteria:} Different datasets might have similar labels but with varied labeling criteria. This could account for the observed discrepancies in our results, a sentiment discussed in prior research\cite{zhao2023chatgpt}.
    
    \item \textbf{Potential for Further Optimization:} It is noteworthy that our experimental results might not fully reflect the optimal performance of ChatGPT for the corresponding tasks. As our results underscore the model's sensitivity to specific label texts, an adjustment of prompt could enhance its performance. We believe that through more refined prompt engineering and careful selection of context examples, the performance of ChatGPT can be further enhanced.

    \item \textbf{Model Specificity:} Our study focused exclusively on the evaluation of ChatGPT, a representative large language model. Whether similar results would be observed when using other models remains undetermined. 
\end{enumerate}

In future research, a broader study that encompasses various linguistic and cultural contexts, as well as a wider range of emotion labels and models, could offer a more holistic perspective on LLMs' strengths and shortcomings in emotion recognition.

\begin{figure}[ht]
\begin{mdframed}[backgroundcolor=black]
\begin{flushleft}
\color[rgb]{0.8,0.8,0.8}\scriptsize
(User) In the context of emotion recognition research, what are the commonly used emotion labels?

\color[rgb]{0.8,0.8,0.8}\scriptsize
(ChatGPT) In emotion recognition research, there are several commonly used emotion labels that researchers use to categorize and study human emotional states. These labels are typically based on various emotion theories and models. Some of the most common emotion labels include:
\begin{enumerate}
   \item Basic Emotions: These are often considered the core, universally recognized emotions proposed by Paul Ekman. They include:
     \begin{itemize}
     \item Happiness
     \item Sadness
     \item Anger
     \item Fear
     \item Disgust
     \item Surprise
     \end{itemize}

   \item Emotion Intensity: Researchers may also measure the intensity or strength of emotions, ranging from very weak to very strong.

   \item Valence-Arousal Model: This model categorizes emotions based on their valence (positive, negative, or neutral) and arousal (intensity or activation). Common labels include:
     \begin{itemize}
     \item Positive emotions (e.g., joy, love)
     \item Negative emotions (e.g., anger, sadness)
     \item Neutral emotions (e.g., boredom, indifference)
     \end{itemize}

   \item Plutchik's Wheel of Emotions: Robert Plutchik's model suggests that there are eight primary emotions, which can be combined to form more complex emotional states. These primary emotions include:
     \begin{itemize}
     \item Joy
     \item Sadness
     \item Anger
     \item Fear
     \item Trust
     \item Disgust
     \item Surprise
     \item Anticipation
     \end{itemize}
\end{enumerate}
...
It's important to note that the choice of emotion labels can vary depending on the specific goals of the research and the theoretical framework being used. Additionally, advances in emotion recognition technology and machine learning have led to more fine-grained and automated approaches for detecting and categorizing emotions, which may use a combination of these labels or even entirely different methods.
\end{flushleft}
\end{mdframed}
\caption{
An example of the output of ChatGPT regarding commonly used emotion labels
}
\label{fig:chatgpt_response}
\end{figure}

\section{Conclusion}
Our exploration of ChatGPT's capabilities in emotion recognition has illuminated both its potential and pitfalls. While the model demonstrates impressive proficiency, especially when fine-tuned, it is susceptible to inconsistencies across various emotion labels and datasets. These variations might be influenced by underlying biases, highlighting the importance of meticulous dataset and label selection. As the allure of deploying models like ChatGPT in critical domains grows, it is paramount to approach their integration with caution and diligence, ensuring their application remains reliable and unbiased in real-world contexts.

\section*{Acknowledgment}
This study was conceptualized, conducted, and written by the authors, and an AI (OpenAI's GPT-4 model) was used for proofreading.

\bibliographystyle{ieeetr}
\bibliography{bib}

\end{document}